%% file: main.tex
\documentclass[a4paper,fleqn]{cas-sc}
\usepackage[numbers]{natbib}
\usepackage{float}
\usepackage{placeins}
\usepackage{tabularx}
\usepackage{booktabs}
\usepackage{array}
\usepackage{graphicx}

\def\tsc#1{\csdef{#1}{\textsc{\lowercase{#1}}\xspace}}
\tsc{WGM}
\tsc{QE}
\tsc{EP}
\tsc{PMS}
\tsc{BEC}
\tsc{DE}

\begin{document}

\makeatletter
\def\printorcid{}
\makeatother

\shorttitle{3D-PRIMME for 3D grain growth dynamics}
\shortauthors{Tian et~al.}

\title[mode=title]{A Physics-Regulated Neural Framework for Learning 3D Grain Growth Dynamics}    

\author[1]{Zhihui Tian} \cormark[1] \ead{zhihui.tian@ufl.edu}
\author[1]{Kang Yang}
\author[2]{Michael Tonks}
\author[3]{Amanda R. Krause}
\author[1]{Joel B. Harley} \cormark[1] \ead{joel.harley@ufl.edu}

\cortext[cor1]{Corresponding authors}
\affiliation[1]{
organization={Department of Electrical and Computer Engineering, University of Florida},
city={Gainesville},
state={FL},
country={USA}
}

\affiliation[2]{
organization={Department of Materials Science Engineering, University of Florida},
city={Gainesville},
state={FL},
country={USA}
}

\affiliation[3]{
organization={Department of Materials Science Engineering, University of Florida, Carnegie Mellon University},
city={Pittsburgh},
state={PA},
country={USA}
}
\begin{abstract}
Grain growth is governed by the reduction in grain boundary energy and exhibits well-established statistical scaling laws. Developing data-driven surrogates that preserve these physical invariants while remaining computationally scalable remains challenging, especially in 3D. We present 3D-PRIMME (Physics-Regulated Interpretable Machine Learning for Microstructure Evolution) for learning three-dimensional grain growth dynamics. The model is trained using only two consecutive time steps yet accurately reproduces the linear coarsening law and preserves topological statistics over extended time scales. Despite being trained on a 
$100^3$ grid points with 512 grains, the learned evolution operator is applied to domains up to $1024^3$ grid points with 550000 grains without retraining, maintaining consistent kinetics and grain topology across orders-of-magnitude increases in system size. These results demonstrate that 3D-PRIMME learns a scale-independent and temporally stable local evolution rule, enabling efficient and robust large-scale surrogate prediction of 3D microstructure evolution.
\end{abstract}




\maketitle

\section{Introduction}

Grain growth is a fundamental microstructural process that plays a central role in determining the macroscopic properties of polycrystalline materials, including mechanical strength and corrosion resistance. During thermal treatments, grain growth governs the evolution of grain size, morphology, and topology, shaping the statistical characteristics of polycrystalline microstructures. Understanding and predicting microstructure evolution is therefore a longstanding and fundamental challenge in materials science, particularly due to its inherently multiscale nature in both space and time. Physics-based simulation methods, such as kinetic Monte Carlo \cite{wu1982potts} and phase-field models \cite{chen2002phase}, have been widely used to study microstructure evolution due to their strong physical interpretability and predictive capability. Other computational approaches, such as cellular automata \cite{he2006computer}, front tracking \cite{lazar2011more}, and level set methods \cite{fausty2021new} have also
been applied to model grain growth. These approaches explicitly model the underlying thermodynamic and kinetic mechanisms governing microstructural changes, and have provided valuable insights into grain growth behavior and statistical laws. However, accurately resolving microstructure evolution over large spatial domains and long times typically requires substantial computational resources, which limits their scalability and practical applicability. Moreover, although physics-based models such as Monte Carlo Potts and phase-field simulations capture many essential mechanisms of microstructure evolution, they often involve idealized assumptions and simplified material descriptions, which may limit their ability to fully reproduce the complexity of real experimental microstructures \cite{miodownik2002review}. This motivates the development of machine learning approaches that can learn directly from experimental data.

In recent years, data-driven approaches based on machine learning have recently been explored as surrogate models for microstructure evolution \cite{yan2022novel,fan2024accelerate,yang2021self, peivaste2025teaching,tep2025high,tian2025scaling}. By learning evolution patterns directly from simulation data, these methods offer the potential to significantly accelerate predictions while retaining essential microstructural features. Although deep learning has been increasingly applied to modeling microstructure evolution, relatively few studies have focused on 3D datasets. Meanwhile, advances in 3D characterization techniques are enabling experimental measurement and reconstruction of 3D microstructures~\cite{ludwig2009new,holm2006three}. Developing learning frameworks capable of handling 3D data will help leverage these emerging experimental capabilities. However, extending existing methods from 2D to 3D remains challenging because grain growth is inherently a spatiotemporal problem: the microstructure evolves in three spatial dimensions and one temporal dimension. Many existing learning approaches operate on the full microstructure field at each time step, leading to a dramatic increase in memory requirements. Among these, only graph neural networks have shows successful results in 3D, but have a large computational cost \cite{fan2024accelerate}. 

To address this problem, the microstructure can be first encoded with a neural network into a latent space, and the dynamical evolution is then modeled in that latent space \cite{tian2025scaling}. However, this also requires a large computational and memory cost, as the potentially large domains need to be input into the network and encoded for each global microstructure. Yet, grain growth dynamics are largely governed by local interactions at grain boundaries, where the evolution of an interface point primarily depends on its local neighborhood. Classical models often attribute this behavior to curvature-driven grain boundary migration, although other local energetic mechanisms may also contribute~\cite{yamakov2006relation,xu2024grain}. This locality suggests that accurate prediction of microstructure evolution can be achieved without explicitly encoding the entire global system. Motivated by this observation, we instead learn a local evolution rule that updates the microstructure through neighborhood-based interactions. By operating on local patches rather than the full domain, the proposed framework avoids the memory bottleneck associated with large 3D systems while preserving the essential physical mechanisms of grain boundary motion.

In this work, we present a deep learning framework for modeling grain growth dynamics in 3D spatial domains. Specifically, we extend our previous PRIMME model from 2D to 3D through targeted modifications to the network architecture, physics regularization, and loss function~\cite{yan2022novel}. The present approach enables accurate prediction of very large 3D grain growth datasets on voxel grids up to $1024^3$ grid points over long times, demonstrating strong capability in both spatial and temporal extrapolation. In addition, the model is operates with minimal supervision, enabling effective learning of grain growth dynamics from limited training data. We also investigate the uncertainty in the model predictions. Model validation is conducted across multiple datasets, and the effects of hyperparameters and training data on model performance are systematically analyzed.

\section{Methodology}
3D-PRIMME is designed to learn the local evolution rules governing 3D microstructure dynamics from data. Rather than relying on a global representation of the microstructure, the framework operates on fixed-size local spatial windows, which makes it naturally scalable to large 3D domains. For each target site, the local neighborhood is not represented directly by the original grain IDs, since these IDs are arbitrary and carry no physical meaning. Instead, following the Hamiltonian formulation~\cite{mason2015kinetics,anderson1984computer}, the neighborhood configuration is mapped to an interface-site representation. This structural metric counts neighboring sites with different grain IDs within the observation window, thereby encoding the local grain-boundary environment that serves as the model input. The corresponding supervision is given by the site label in the subsequent microstructure state, so that the model learns the updated rule from the observed microstructure evolution. In the following subsections, we describe the generation of training data, the architecture and hyperparameters, the validation metrics, and experimental setup.

\subsection{Training data from stochastic model}
The training and evaluation datasets are generated using the mode-filter (MF) grain growth model~\cite{melville2024new}. 
The MF model is adopted for three main reasons. First, the stochasticity inherent in the MF update rule introduces fluctuations in the evolving microstructures, which provide a useful proxy for the noise commonly observed in experimental datasets. Second, the grain growth kinetics can be systematically controlled through the choice of neighborhood kernel, enabling the simulation of different growth rates and inclination-dependent grain boundary behavior. Finally, it is computationally efficient, allowing rapid generation of 3D simulation data. Here, we give a brief summary of the MF model but see Melville et al.~\cite{melville2024new} for more information.

In the MF framework, microstructure evolution is modeled through repeated local updates based on the most frequent orientation within a predefined neighborhood, which provides an efficient approximation of curvature-driven grain growth. It is theoretically similar to the Monte Carlo Potts model, but takes advantage of the MF algorithm to simplify the numerical implementation. The MF processes each site independently by sampling neighboring sites using a designed kernel. In the MF model, the updated grain state of a target site is selected by minimizing the local Hamiltonian energy:
\begin{equation}
s_* = \arg\min_s w_s, \qquad \mathbf{w}=\hat{\Gamma}\mathbf{c}+\mathbf{u},
\end{equation}
where \(s_*\) is the updated grain state, \(\mathbf{w}\) contains the local energy associated with each candidate grain state, \(\hat{\Gamma}\) is the grain-boundary energy matrix, \(\mathbf{u}\) is a bulk energy term, and \(\mathbf{c}\) represents the local neighborhood composition.

The neighborhood composition \(\mathbf{c}\) is computed from a Gaussian-sampled local neighborhood. Specifically, the weighted count of grain state \(j\) can be written as
\begin{equation}
c_j = \sum_{k \in \mathcal{N}} f(\Delta_k)\,\mathbb{I}(s_k=j),
\end{equation}
where \(\Delta\) is the displacement vector from the target site to neighboring site \(k\), \(s_k\) is the grain state at that neighbor, and \(f(\Delta)\) gives the sampling weight. The Gaussian sampling distribution is
\begin{equation}
f(\Delta)=
\frac{1}{(2\pi)^{d/2}|\Sigma|^{1/2}}
\exp\left(
-\frac{1}{2}\Delta^\mathrm{T}\Sigma^{-1}\Delta
\right),
\end{equation}
where \(\Sigma\) is the covariance matrix controlling the shape and orientation of the sampled neighborhood. The grain growth is isotropic when the covariance matrix \(\Sigma = a \mathbf{I}\), where \(a\) is the variance. By using an anisotropic covariance matrix,
the weighted neighborhood composition \(\mathbf{c}\) becomes direction dependent, and the same MF energy-minimization rule produces inclination-dependent grain-boundary migration. Inclination-dependent training data were generated using this approach with the MF model.

We run 200 isotropic grain growth simulations using the 3D mode-filter model, producing 200 sequences of evolving grain structures. Each sequence starts from a different initial grain structure containing 512 grains created using a Voronoi tesselation represented on a volume of \(100^3\) grid points and evolves for 100 time steps, resulting in a dataset of shape (200,100,100,100,100), corresponding to (sequence, time, x, y, z). From this dataset, M sequences and 2 consecutive time steps are sampled to construct different training sets, yielding datasets of shape (M, 2, 100, 100, 100), where M is a hyperparameter that controls the amount of training data. Smaller values of M correspond to more data-limited settings that are common with experimental data. 

We construct a similar data set of 200 anisotropic simulations with 
\begin{equation}
    \Sigma = \begin{pmatrix}
    a & b & b\\
    b & a & b\\
    b & b & a
    \end{pmatrix},
\end{equation} 
where $b$ is the scalar covariance, and use \(a=25\) and \(b=20\). As a result, neighbor sampling becomes directionally biased, leading to anisotropic grain growth along \(\mathbf{u}_{111}=(1,1,1)/\sqrt{3}\), corresponding to the \([111]\) direction.

\subsection{Architecture and Hyperparameters}

\begin{figure}[pos=htbp]
  \centering
\includegraphics[width=\linewidth]{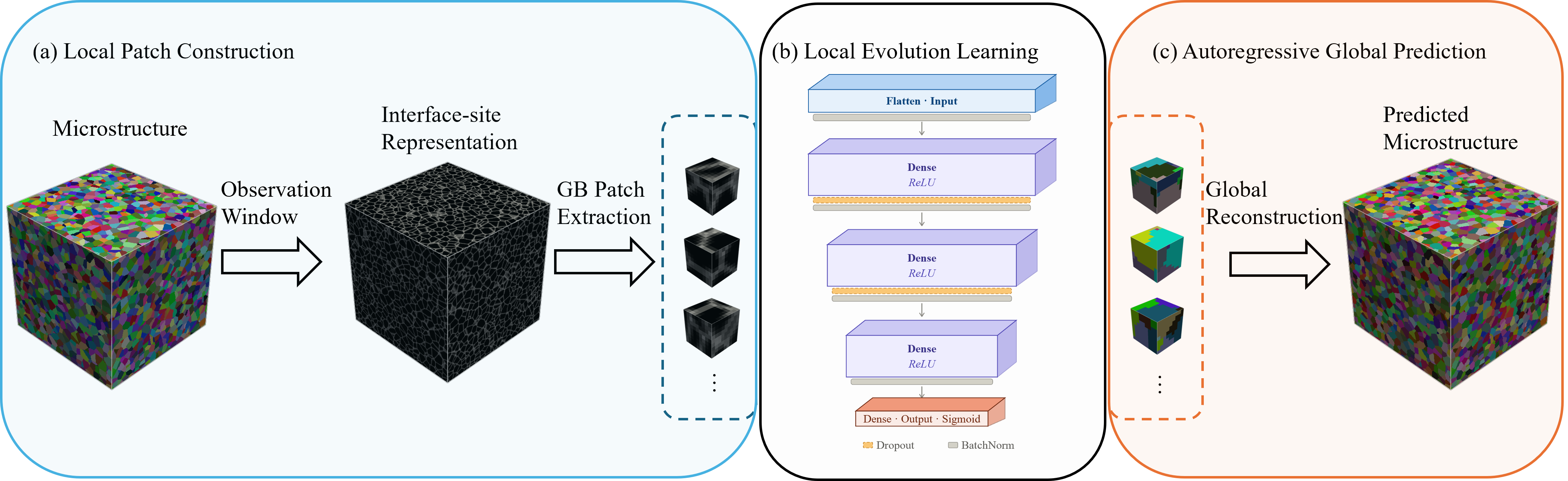}
  \caption{Workflow of the proposed framework for learning and predicting 3D microstructure evolution.
(a) The input 3D microstructure is first transformed into an interface-site representation, from which local grain-boundary patches are extracted to construct training samples. 
(b) A neural network is trained to map each local patch to its evolved state, enabling parallel optimization of local microstructure updates. 
(c) During inference, the trained model is applied autoregressively to predict local block evolution, and the predicted blocks are aggregated to reconstruct the global 3D microstructure.}
\label{fig:architecture}
\end{figure}

The overall architecture of the 3D-PRIMME framework is illustrated in Fig.~\ref{fig:architecture}. The method follows a divide-and-conquer strategy to enable scalable simulation of large 3D microstructures. Specifically, the global microstructure is first encoded into interface-site representation and then divided into local patches for neural network. During training, the local features from the current microstructure together with the corresponding local labels derived from the next microstructure are used to train a neural network that predicts local updates. During inference, the trained network takes only the current local features as input and predicts the corresponding local evolution. These local updates are then combined to reconstruct the global microstructure at the next time step. Since the updates are computed independently within local windows, the framework enables highly parallel computation and significantly reduces memory requirements for large-scale 3D simulations. Detail of parameter settings are provided in Table ~\ref{tab:primme_components}.

\begin{table}[htbp]
\centering
\renewcommand{\arraystretch}{1.18}
\caption{
Summary of key components and tensor shapes in 3D-PRIMME framework.
$N_o$ and $N_a$ denote the observation and action window sizes, respectively.
$M$ denotes the number of boundary-centered sites extracted from the microstructure, and
$\mathcal{A} \in \mathbb{R}^{M \times N_a \times N_a \times N_a}$ denotes the collection of action-window tensors.
}
\label{tab:primme_components}
\begin{tabularx}{\linewidth}{
    >{\raggedright\arraybackslash}p{0.24\linewidth}
    >{\raggedright\arraybackslash}X
    >{\raggedright\arraybackslash}p{0.18\linewidth}
}
\toprule
\textbf{Component} & \textbf{Description} & \textbf{Shape} \\
\midrule

Microstructure
& 3D grain-ID field representing the evolving microstructure
& $N \times N \times N$ \\
\addlinespace[0.25em]

Observation window
& Defines the local neighborhood used to compute the interface-site representation
& $N_o \times N_o \times N_o$ \\
\addlinespace[0.25em]

Interface-site representation
& Voxel-wise structural metric that encodes the local number of interface sites within the observation window
& $N \times N \times N$ \\
\addlinespace[0.35em]

Action window
& Extracts boundary-centered local neighborhoods from the interface-site representation
& $N_a \times N_a \times N_a$ \\
\addlinespace[0.25em]

Local features
& Neural-network input features extracted from the action-window tensor
& $\mathcal{A} \in \mathbb{R}^{M \times N_a \times N_a \times N_a}$ \\
\addlinespace[0.25em]

Local labels
& Ground-truth state-flip indicators within the action window
& $\mathcal{A}$ \\
\addlinespace[0.25em]

Neural network
& Learns the mapping from local features to local state-flip probabilities using the architecture from Ref.~\cite{yan2022novel}
& $\mathcal{A} \rightarrow \mathcal{A}$ \\
\addlinespace[0.25em]

Local update probabilities & Predicted probabilities of candidate state flips within the action window & $\mathcal{A}$ \\ \addlinespace[0.25em] 

Predicted microstructure & Global microstructure reconstructed by assembling the predicted local updates & $N \times N \times N$ \\

\bottomrule
\end{tabularx}
\end{table}

To construct the model input, we first define the interface-site representation at site $i$ and time $t$ as
\begin{equation}
S_i^{(t)}
=
\sum_{j \in \mathcal{N}_{\mathrm{o}}(i)}
\left(1-\delta_{ij}^{(t)}\right),
\end{equation}
where $\mathcal{N}_{\mathrm{o}}(i)$ denotes the observation window centered at site $i$, $s_i^{(t)}$ is the grain label at site $i$, and $\delta_{ij}^{(t)}$ denotes the Kronecker delta comparing the grain labels of sites $i$ and $j$ at time $t$. This quantity counts the number of sites within the observation window that belong to grains different from the target site, thereby providing a local structural metric of the grain-boundary environment around site $i$.

The training target is constructed by checking whether site $i$ adopts the current grain label of a candidate site $j$ in the action window after one evolution step. For each candidate site $j \in \mathcal{N}_{\mathrm{a}}(i)$, we define the binary label as
\begin{equation}
y_{ij}^{(t)}
=
\begin{cases}
1, & s_i^{(t+1)} = s_j^{(t)},\\
0, & s_i^{(t+1)} \neq s_j^{(t)}.
\end{cases}
\end{equation}
This binary label indicates whether site $i$ adopts the current grain label of neighbor $j$ at the next time step.

A squared-error loss is used to link the interface-site inputs to the predicted action likelihoods. The loss for site $i$ at time step $t$ is defined as
\begin{equation}
L_i^{(t)}
=
\frac{1}{|\mathcal{N}_{\mathrm{a}}(i)|}
\sum_{j \in \mathcal{N}_{\mathrm{a}}(i)}
\left|
Y_\theta\left(S_i^{(t)},S_j^{(t)}\right)
-
y_{ij}^{(t)}
\right|^2 ,
\end{equation}

where $\mathcal{N}_{\mathrm{a}}(i)$ denotes the action window centered at site $i$, $Y_\theta(\cdot)$ is the neural-network prediction, and $S_i^{(t)}$ and $S_j^{(t)}$ are the interface-site representations at sites $i$ and $j$, respectively. The original PRIMME model~\cite{yan2022novel} included a regularization term in the loss function to penalize increases in the number of grain-boundary sites. In 3D-PRIMME, this explicit regularization term is removed. Instead, the interface-site representation, together with the local observation and action windows, constrains the learned update to the local grain-boundary environment. This point is discussed further in Section~\ref{sec:discussion}.

The division of the global microstructure into small local windows is governed by two physically interpretable design parameters: the observation window size $N_o$ and the action window size $N_a$. These two windows act as complementary receptive fields at different stages of the framework. The observation window size $N_o$ defines the local neighborhood used to transform the original grain-ID microstructure into the interface-site representation. Because this transformation is performed before neural-network prediction, $N_o$ can be selected based on the characteristic length scale of the data, such as the average grain size, to ensure that the interface-site representation captures the relevant local structural context. In contrast, the action window size $N_a$ defines the local patch extracted from the interface-site representation and therefore determines the input size of the neural network. This separation provides a flexible way to control the local microstructural context used to construct the interface-site metric independently from the size of the neural-network input. As a result, the model can adapt to different microstructural length scales while maintaining a compact and consistent learning architecture. The sensitivity analysis in Fig.~\ref{fig:tuning} further shows that the model behavior depends on these two window sizes, indicating that these receptive fields should be informed by characteristic microstructural length scales rather than treated as universal constants. 


\subsection{Validation Metrics}

To evaluate the performance of the proposed framework, we employ statistical validation protocols based on both physical metrics and voxel-level accuracy. The statistical analysis of the long-term predicted microstructure evolution evaluates whether the predictions remain consistent with expected grain-growth behavior over time. Specifically, several key grain growth metrics are examined during inference. First, the evolution of the squared average grain radius $\left<r\right>^2$ is analyzed to verify the expected linear $\left<r\right>^2$–time scaling of curvature-driven grain growth\cite{wakai2000three}. Second, the evolution of the average number of grain faces is monitored to assess the preservation of topological characteristics. In 3D isotropic grain growth, the average number of grain faces is expected to converge to approximately 13 to 14 \cite{kamachali20123}. To further examine the coupling between geometry and topology, the relationship between $\left<r\right>^2$ and the number of grain faces is also evaluated. In addition to these statistical measures, voxel-based prediction accuracy is reported to quantify the agreement between the predicted and reference microstructures at the voxel level. The voxel-wise accuracy at time step $t$ was computed as:
\begin{equation}
\mathrm{Acc}(t)=\frac{1}{N}\sum_{i=1}^{N}\mathbf{1}\!\left[\hat{y}_i^{(t)} = y_i^{(t)}\right], \label{eq:voxel-wize_accuracy}
\end{equation}
where $N$ is the total number of voxels, and $\hat{y}_i^{(t)}$ and $y_i^{(t)}$ denote the predicted and ground-truth grain labels of voxel $i$ at time step $t$, respectively. Finally, 3D visualizations of the microstructure evolution are provided to qualitatively assess the spatial structure and morphological evolution during the evolution.

\subsection{Implementation Details}

All studies follow a unified simulation setup. The neural network is trained on microstructures generated by the MF model with $100^3$ grid points. Training samples are constructed from short evolution sequences extracted from the MF simulations, with the number of supervised future steps varied according to the specific study. The dataset is divided into training and validation sets for model optimization and hyperparameter selection. During inference, the trained network is applied autoregressively to generate long-term evolution of up to 100 evolution steps. Unless otherwise noted, the same training pipeline, optimizer, and loss function are used throughout to ensure consistency in the comparisons. The network is implemented in PyTorch and trained on a single NVIDIA B200 GPU.
To systematically evaluate the proposed framework, we design a set of test scenarios addressing five key aspects:
\begin{itemize}
    \item \textbf{Receptive-field sensitivity:} We evaluate the influence of the observation and action window sizes, \(N_o\) and \(N_a\), on model predictions. These parameters control the construction of the interface-site representation and the neural-network input size, respectively.

    \item \textbf{Training variability:} We assess the sensitivity of the model to stochastic factors during training, including random initialization and data shuffling. To quantify this variability, models are trained multiple times using identical hyperparameters but different random seeds.

    \item \textbf{Spatial scalability:} We examine whether the learned local update rule can generalize to spatial domains much larger than those used during training. Specifically, models trained on \(100^3\) grid points microstructures are evaluated on larger domains, including \(256^3\), \(512^3\), and \(1024^3\) grid points.

    \item \textbf{Training-data sufficiency:} We investigate how much training data and temporal supervision are required for the model to learn physically consistent grain-growth dynamics. 

    \item \textbf{Inclination-dependent grain growth:} We test whether 3D-PRIMME can learn inclination-dependent grain-growth behavior from anisotropic MF data without explicitly providing boundary inclination as an input feature. This study evaluates the ability of the model to infer direction-dependent boundary migration behavior from the local microstructural evolution.
\end{itemize}

\section{Results}

\subsection{Sensitivity to observation and action window sizes}

\begin{figure}[pos=htbp]
  \centering
  \includegraphics[width=0.8\linewidth]{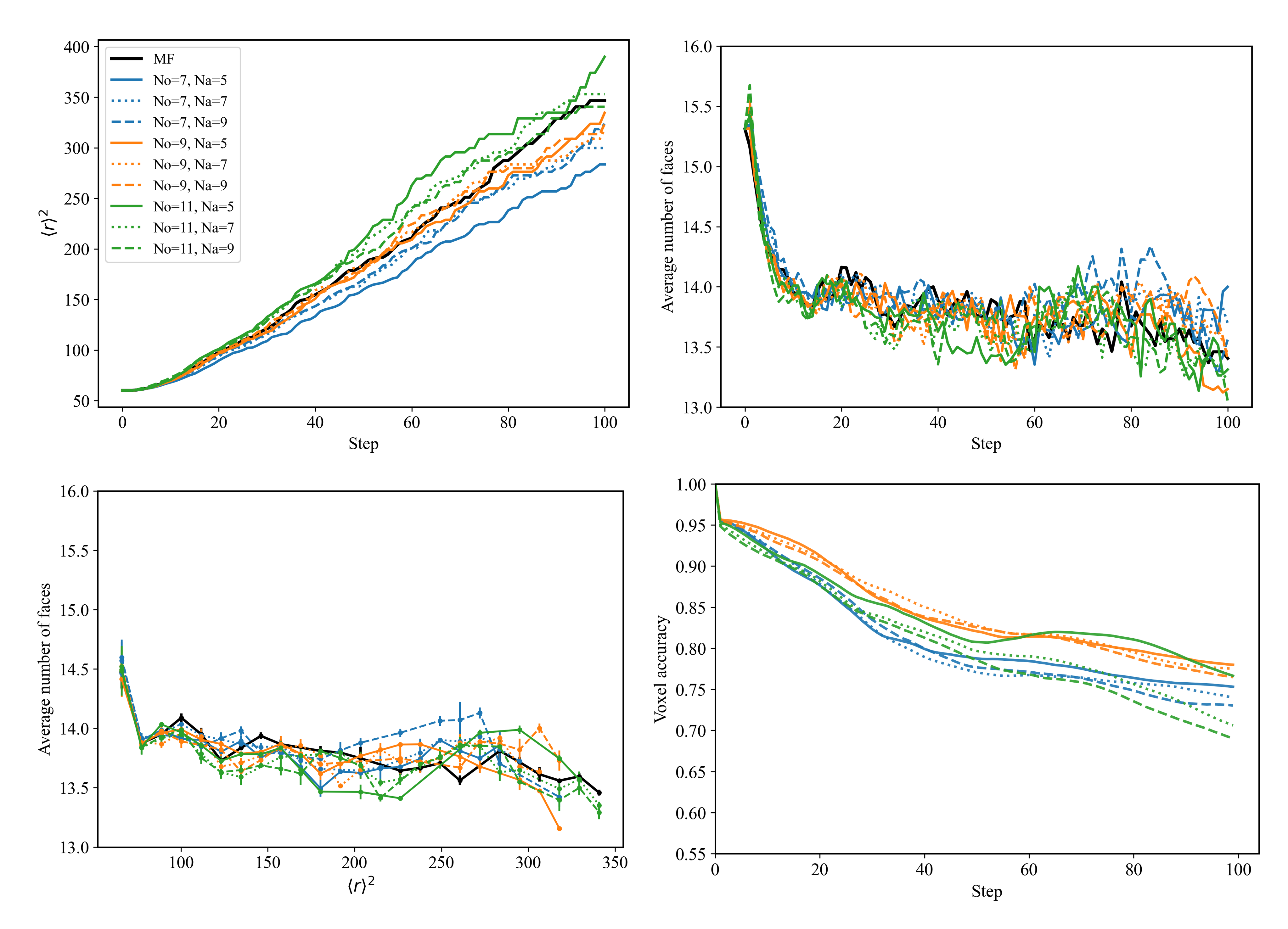}
  \caption{Investigation of the impact of observation and action window sizes ($N_o$ and $N_a$, respectively) on the 3D-PRIMME predictions for a $100^3$ grid points with 512 initial grains. (a) Mean squared grain size $\langle r \rangle^2$ versus time step. (b) Average number of grain faces versus time step. (c) Average number of grain faces versus $\langle r \rangle^2$. (d) Voxel-wise accuracy computed using Eq.~\ref{eq:voxel-wize_accuracy} versus time step. A MF result is shown for reference in (a)-(c).}
\label{fig:tuning}
\end{figure}

We begin by evaluating the predictive performance of 3D-PRIMME under different observation and action window sizes in a volume of $100^3$ grid points. All cases start with 512 initial grains with the same initial grain structure created with a Voronoi tesselation. The model is trained using one sequence of two consecutive times steps. The change of the squared average grain radius $\left<r\right>^2$ with number of time steps (Fig.~\ref{fig:tuning}a) shows that all model configurations reproduce the approximately linear coarsening kinetics observed in the MF training data. However, the slope of the line, or the growth rate, can vary with the window sizes.The observation window size has a strong influence on the growth rate: smaller observation windows tend to underpredict the MF kinetics, whereas larger observation windows tend to accelerate coarsening. In contrast, the action window size, $N_a$, has a comparatively weaker and less systematic effect on the growth rate. An observation window of $N_o=9$ results in the most accurate reproduction of the MF growth rate. 

The relationship between the average number of grain faces with number of time steps  (Fig.~\ref{fig:tuning}b) also remains fairly consistent with the MF results, indicating that the learned dynamics preserve the key topological characteristics of grain growth. Again, $N_o=9$ provides the best overall performance, whereas no clear trend is observed with respect to the action window size $N_a$. 
This suggests that the observation window has a stronger influence on the learned dynamics because it directly determines the local neighborhood used to construct the interface-site representation. 
An insufficient $N_o$ may fail to capture the relevant grain-boundary environment, while an overly large $N_o$ may introduce redundant information and increase the learning complexity. 
By comparison, $N_a$ only controls the size of the local feature patch used for prediction. 
Once the relevant physical context is encoded through $N_o$, increasing $N_a$ does not necessarily improve the prediction, leading to less dependence on the action window size. Similar behavior is seen in the relationship between the average number of grain faces and $\left<r\right>^2$ (Fig.~\ref{fig:tuning}c). 

Finally, we directly compare the voxel-wise accuracy for the various window sizes (Fig.~\ref{fig:tuning}d) calculated using Eq.~\ref{eq:voxel-wize_accuracy}, which decreases with number of time steps in all cases. This reduction does not necessarily indicate incorrect physical behavior; rather, it reflects the increasing divergence away from the initial condition between the stochastic prediction of the microstructure by MF and the deterministic prediction by the 3D-PRIMME models. Recent ensemble simulations of polycrystalline grain growth have similarly shown that abrupt topological events, such as grain disappearance, can introduce uncertainty and limit trajectory-level predictability in microstructure evolution simulations~\cite{lyu2026limits}. However, high voxel accuracy does indicate a more accurate reproduction of the training data. The voxel-wise accuracy also varies with the window sizes, although its trend does not directly follow the kinetic error measured from $\langle r\rangle^2$. A quantitative comparison further supports the observation from Fig.~\ref{fig:tuning} that $N_o$ has a stronger effect than $N_a$ on the long-term coarsening kinetics. The cases with $N_o=9$ give the lowest relative errors in $\langle r\rangle^2$ compared with the MF reference, ranging from 2.85\% to 3.49\%, while the corresponding errors for $N_o=7$ and $N_o=11$ are generally larger. The setting $N_o=9$ and $N_a=9$ gives the smallest kinetic error, with a relative error of 2.85\%, and is therefore used as the default configuration in the remaining simulations. The effect of $N_a$ is comparatively weaker, although it still affects the average number of grain faces and voxel-wise accuracy. The complete quantitative comparison is provided in Table S1 in the Supporting Information.



\subsection{3D-PRIMME Variability}

\begin{figure}[pos=htbp]
  \centering
\includegraphics[width=0.8\linewidth]{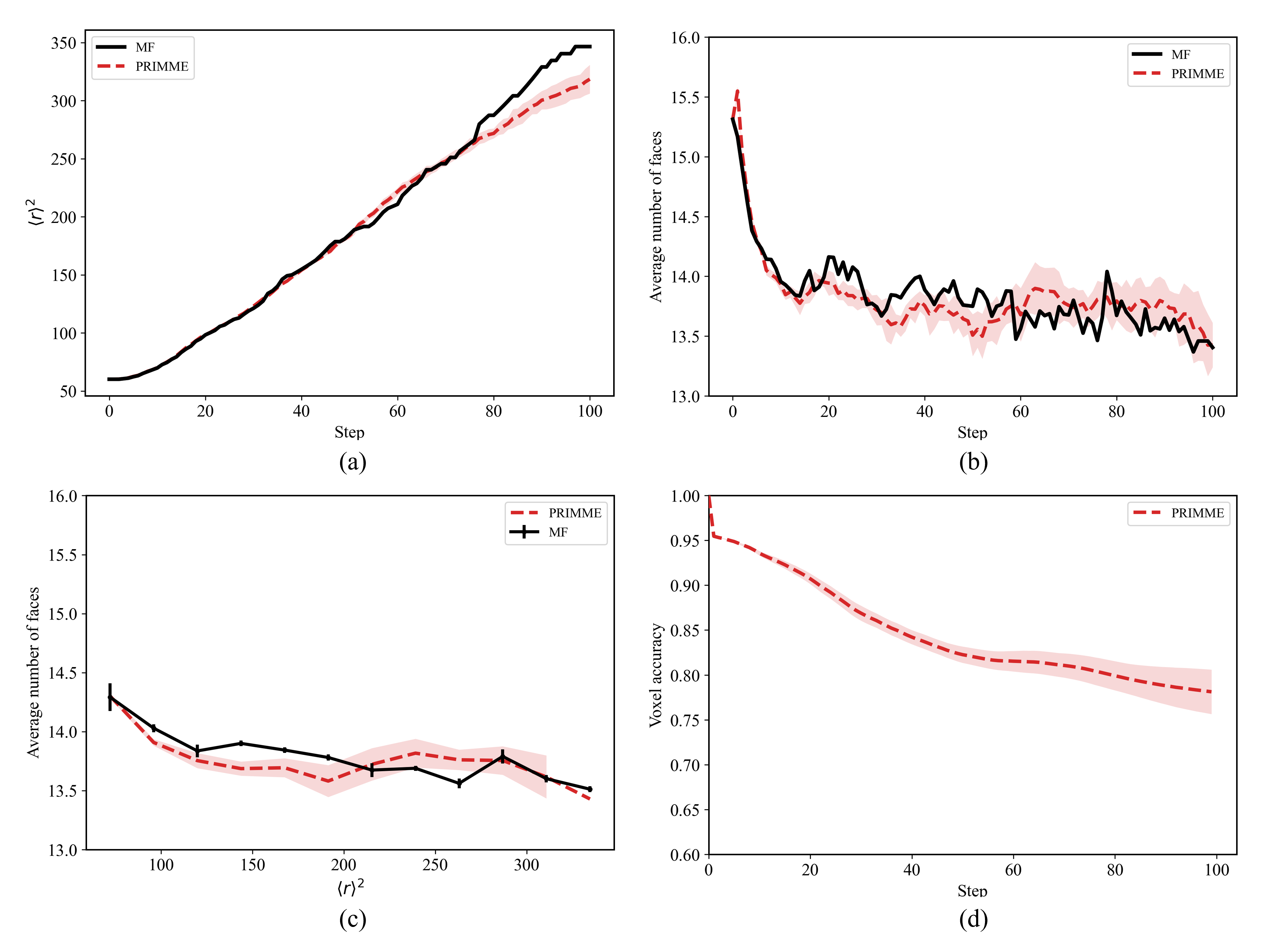}
  \caption{Investigation of the uncertainty in 3D-PRIMME predictions between models trained with the same data and hyperparameters but random initialization and data-shuffling. (a) Mean squared grain size, $\langle r \rangle^2$, versus time step. (b) Average number of grain faces versus time step. (c) Average number of grain faces versus $\langle r \rangle^2$. (d) Voxel-wise accuracy calculated using Eq.~\eqref{eq:voxel-wize_accuracy} versus time step. The solid black curves denote the MF reference data, and the red dashed curves denote the mean 3D-PRIMME predictions. The shaded regions and error bars indicate the corresponding standard deviations across 10 independent models.}
\label{fig:hgyper_tune}
\end{figure}

To assess the uncertainty of the 3D-PRIMME framework, we trained the ten different models using an identical training dataset of one sequence of two consecutive time steps and hyperparameters ($N_o=9$ and $N_a=9$) but different model initialization and data shuffling. We again model grain growth of the same $100^3$ grid points and initial 512 grain structure. The uncertainty is small in the squared average grain radius $\langle r \rangle^2$ versus time, though it increases with time, as shown in Fig.~\ref{fig:hgyper_tune}(a). 
Similarly, uncertainty in the average number of grain faces as a function of time step (Fig.~\ref{fig:hgyper_tune}(b)) and of $\langle r \rangle^2$ (Fig.~\ref{fig:hgyper_tune}(c)) is small. The voxel-wise accuracy computed using Eq.~\eqref{eq:voxel-wize_accuracy} versus time is shown in Fig.~\ref{fig:hgyper_tune}(d), and also exhibits only small uncertainty. These results suggest that the learned representation reliably maintains key statistical invariants of grain growth, despite the non-convex optimization landscape inherent to deep neural networks.

\subsection{3D-PRIMME spatialtemporal scale independence}
Many existing machine learning architectures for grain growth require that training and testing data be trained using the same spatial scales  \cite{fan2024accelerate,yang2021self, peivaste2025teaching,tep2025high,tian2025scaling}. Yet, in 3D grain growth, the computational cost increases rapidly with domain size. Furthermore, prior ultra-large-scale phase-field simulations have shown that sufficiently large spatial and temporal domains are required to retain enough grains for reliable steady-state statistics and to resolve discrepancies in previously reported kinetics and grain-size distributions~\cite{miyoshi2017ultra}. Hence, establishing statistically reliable large-scale grain-growth behavior is challenging for many machine learning architectures. In contrast, PRIMME can be applied to spatialtemporal domains of any size and number of grains without retraining.
 
\begin{figure}[pos=htbp]
  \centering
\includegraphics[width=0.8\linewidth]{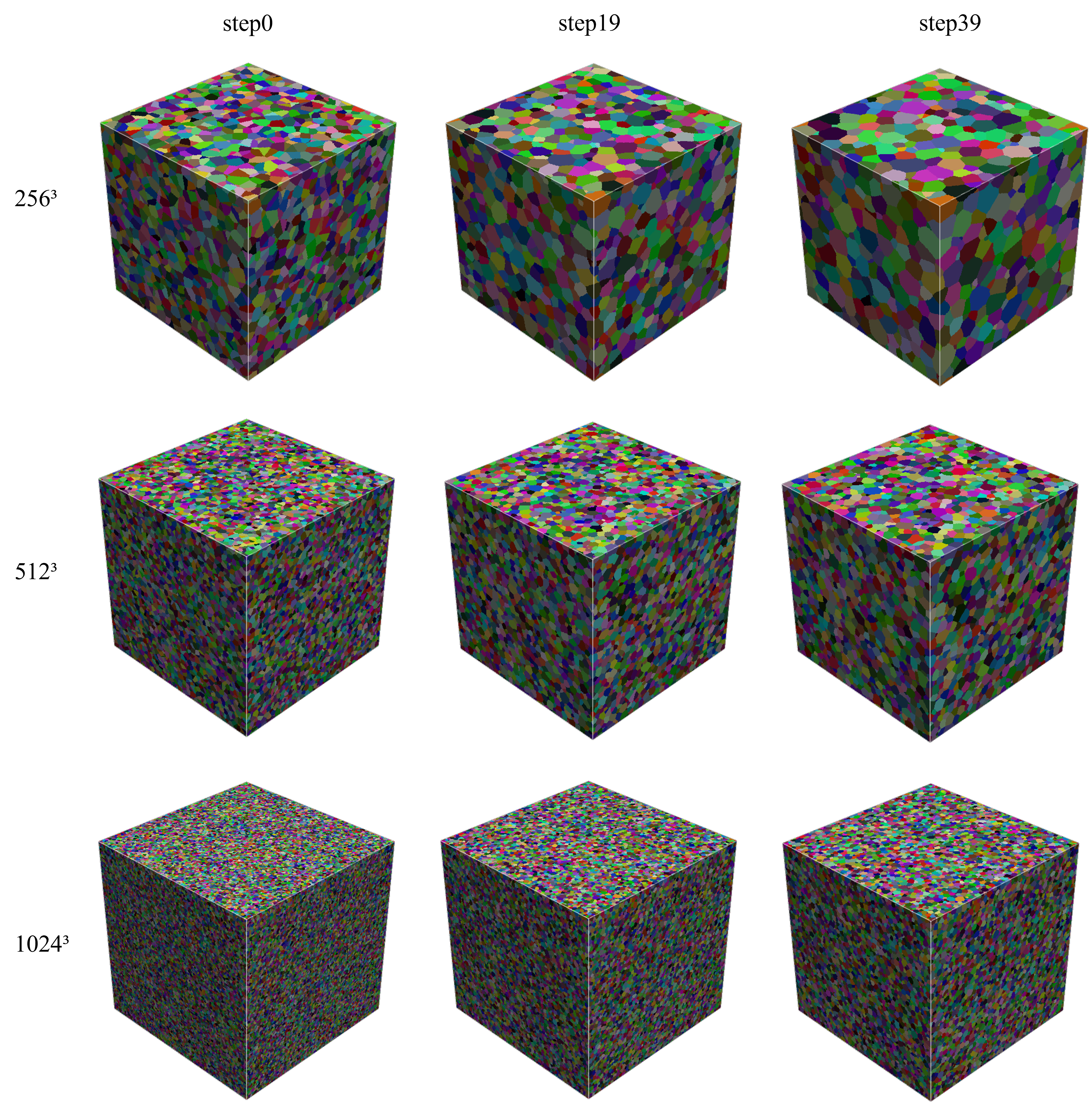}
\caption{
Visualization of 3D-PRIMME grain-growth evolution under extrapolation to larger simulation domains.
Representative snapshots are shown for $256^3$, $512^3$, and $1024^3$ domains at steps 0, 19, and 39.
The three domains contain 8,600, 68,700, and 550,000 initial grains, respectively.
Steps 19 and 39 correspond approximately to representative coarsening states with $N_G \approx N_{G,0}/2$ and $N_G \approx N_{G,0}/4$, respectively, where $N_{G,0}$ is the initial number of grains in each domain.
Different colors denote different grain IDs within each snapshot.
All domains are displayed at the same visual size to facilitate comparison of the grain morphology; the actual domain sizes are indicated by the row labels.
}
\label{fig:spatio temporal}
\end{figure}

We confirm this capability of 3D-PRIMME trained on a single sequence containing only two consecutive time steps from $100^3$ grid points with 512 initial grains to extrapolate to larger domains and more initial grains, such that the initial grain size is equal in each case. We model the grain growth in volumes of $256^3$ grid points with 8,600 initial grains, $512^3$ grid points with 68,700 grains, and $1024^3$ grid points with 550,000 grains, as shown in Fig.~\ref{fig:spatio temporal}. 

\begin{figure}[pos=htbp]
  \centering
\includegraphics[width=0.8\linewidth]{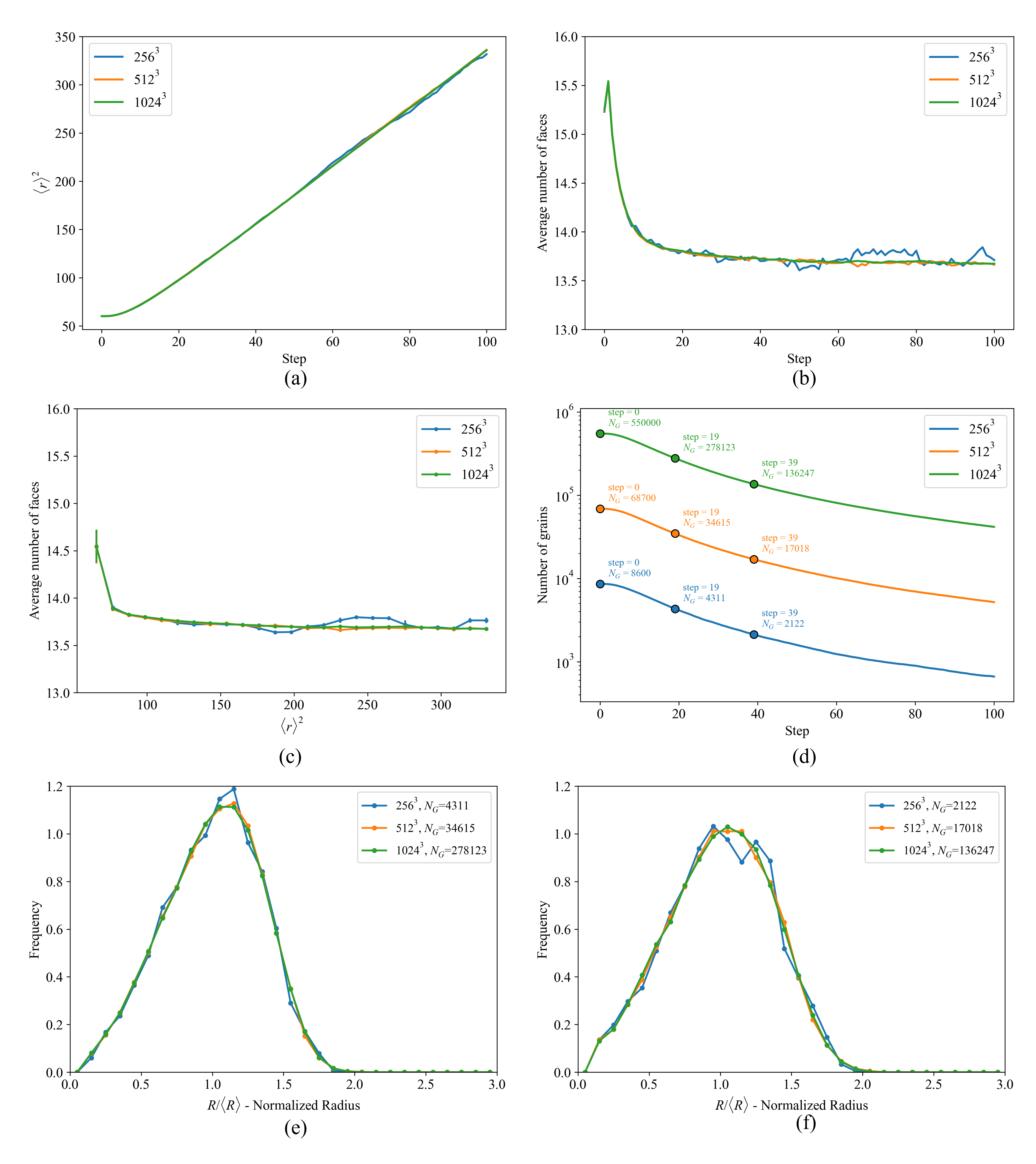}
    \caption{
    Statistical consistency of 3D-PRIMME under extrapolation to larger simulation domains.
    Results are compared for $256^3$, $512^3$, and $1024^3$ domains.
    (a) Evolution of the squared average grain radius, $\langle r\rangle^2$, showing nearly identical coarsening kinetics across different domain sizes.
    (b) Evolution of the average number of grain faces with simulation step.
    (c) Average number of grain faces plotted against $\langle r\rangle^2$, demonstrating collapse of the topological statistics at comparable coarsening states.
    (d) Evolution of the total number of grains, $N_G$, plotted on a logarithmic scale. Markers indicate the representative states selected for cross-size comparison, corresponding approximately to $N_G = N_{G,0}/2$ and $N_G = N_{G,0}/4$, where $N_{G,0}$ is the initial number of grains in each domain.
    (e, f) Normalized grain-radius distributions at representative coarsening states corresponding to $N_G \approx N_{G,0}/2$ and $N_G \approx N_{G,0}/4$, respectively.
    }
\label{fig:spatio temporal statistical}
\end{figure}

The quantitative statistics further confirm this consistency across system sizes. The evolution of the squared average grain radius, $\langle r\rangle^2$, is nearly identical for the three domains, as shown in Fig.~\ref{fig:spatio temporal statistical}(a). After a short initial transient associated with the Voronoi-tessellated initial condition, $\langle r\rangle^2$ evolves approximately linearly with time for all three domain sizes, indicating consistent coarsening kinetics across different system sizes. In addition, the average number of grain faces rapidly approaches a similar steady value in all cases, and its evolution collapses when plotted against $\langle r\rangle^2$ (Fig.~\ref{fig:spatio temporal statistical}(b,c)). The total number of grains, $N_G$, decreases smoothly with time for each domain size (Fig.~\ref{fig:spatio temporal statistical}(d)), and the representative states with $N_G \approx N_{G,0}/2$ and $N_G \approx N_{G,0}/4$ are selected for grain-size distribution comparisons. At these two coarsening states, the normalized grain-radius distributions show strong agreement across different domain sizes (Fig.~\ref{fig:spatio temporal statistical}(e,f)). These results indicate that 3D-PRIMME preserves not only the kinetic scaling behavior, but also the topological evolution and normalized grain-size statistics when extrapolated to larger three-dimensional domains.

Taken together, these results demonstrate that the learned evolution operator is both spatially scale-independent and temporally stable. The present experiments demonstrate stable predictions up to $1024^3$ grid points and 100 evolution steps with 41674(7.6\% of initial number of grains) remaining grains. The model evolves the microstructure through local neighborhood interactions rather than a global representation, which makes extension to substantially larger spatial domains and longer temporal horizons feasible due to its natural suitability for parallel computation. Moreover, by learning local curvature-driven grain boundary migration behavior instead of memorizing domain-specific statistics, the model produces predictions that remain reliable and transferable across domain sizes without retraining.

\subsection{Data-efficient learning of grain growth}

In the previous results, we used a single sequence of two consecutive steps to train 3D-PRIMME. Now, we assess how the increasing the number of sequences influences model performance \cite{hestness2017deep}. For a fair comparison, all models are trained for the same number of parameter-update steps, where each step corresponds to computing the loss from one batch of training samples and updating the neural-network weights once. 
This criterion is used instead of epochs, since one epoch represents a complete pass through the available training data and therefore contains more training samples when more sequences are used. 
Separate 3D-PRIMME models are trained using 1, 10, and 50 sequences that are randomly sampled from the original dataset of 200 simulations. The predictions of these models are assessed by comparing to an independent validation set consisting of 10 MF simulations.

\begin{figure}[pos=htbp]
  \centering
\includegraphics[width=\linewidth]{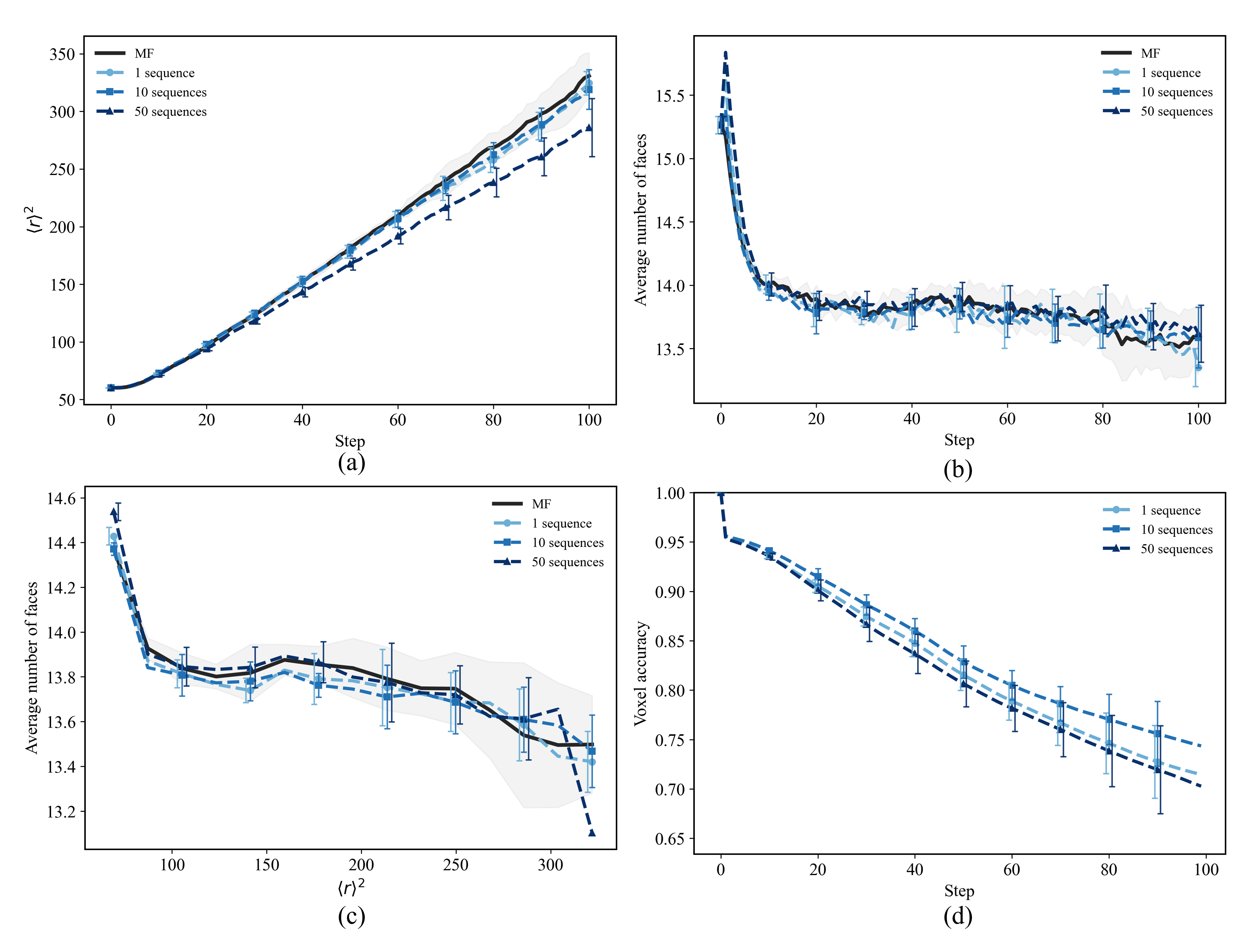}
\caption{Influence of training dataset size on the predicted grain-growth dynamics. 3D-PRIMME models trained with 1, 10, and 50 sequences are shown as light, medium, and dark blue dashed curves, respectively. (a) Evolution of $\langle r \rangle^2$ versus time step. (b) Average number of grain faces versus time step. (c) Average number of grain faces versus $\langle r \rangle^2$. (d) Voxel-wise accuracy versus time step. Each PRIMME curve denotes the mean prediction over 10 independently trained models, with capped error bars indicating the standard deviation across models (horizontally offset between conditions for legibility). The MF results are shown for reference as a solid black line (mean) with a shaded region (standard deviation); MF is omitted in (d) as the voxel accuracy is unity by definition.}
\label{fig:diff_sequence}
\end{figure}

Figure~\ref{fig:diff_sequence} shows the influence of training dataset size on the predicted grain-growth dynamics. As shown in Fig.~\ref{fig:diff_sequence}(a), the evolution of $\langle r\rangle^2$ is linear for all three training data sizes. The growth rate is similar for 1 and 10 sequences, but it slows with 50 sequences. The standard deviation is also larger with 50 sequences. The predicted evolution of the average number of grain faces, shown in Fig.~\ref{fig:diff_sequence}(b), is similar irrespective of the number of sequences. Finally, the voxel accuracy (Fig.~\ref{fig:diff_sequence}(c)) is highest with 10 sequences and has the smallest standard deviation. 50 sequences results in the lowest accuracy and the largest standard deviation.   
A possible explanation for why an increase in training data decreases the accuracy is that the additional training sequences may introduce more variability or redundancy without being fully exploited during optimization \cite{mindermann2022prioritized}. 

Taken together, the above results demonstrate that the proposed framework operates effectively under a minimal data regime. The number of training sequences primarily influences statistical stability but does not alter the learned coarsening laws. These findings indicate that grain-growth dynamics at the considered scale are strongly governed by local behavior, allowing accurate surrogate prediction with limited temporal and sequence-level supervision.


\subsection{Inclination-dependent grain growth}
\begin{figure}[pos=htbp]
  \centering
    \includegraphics[width=\linewidth]{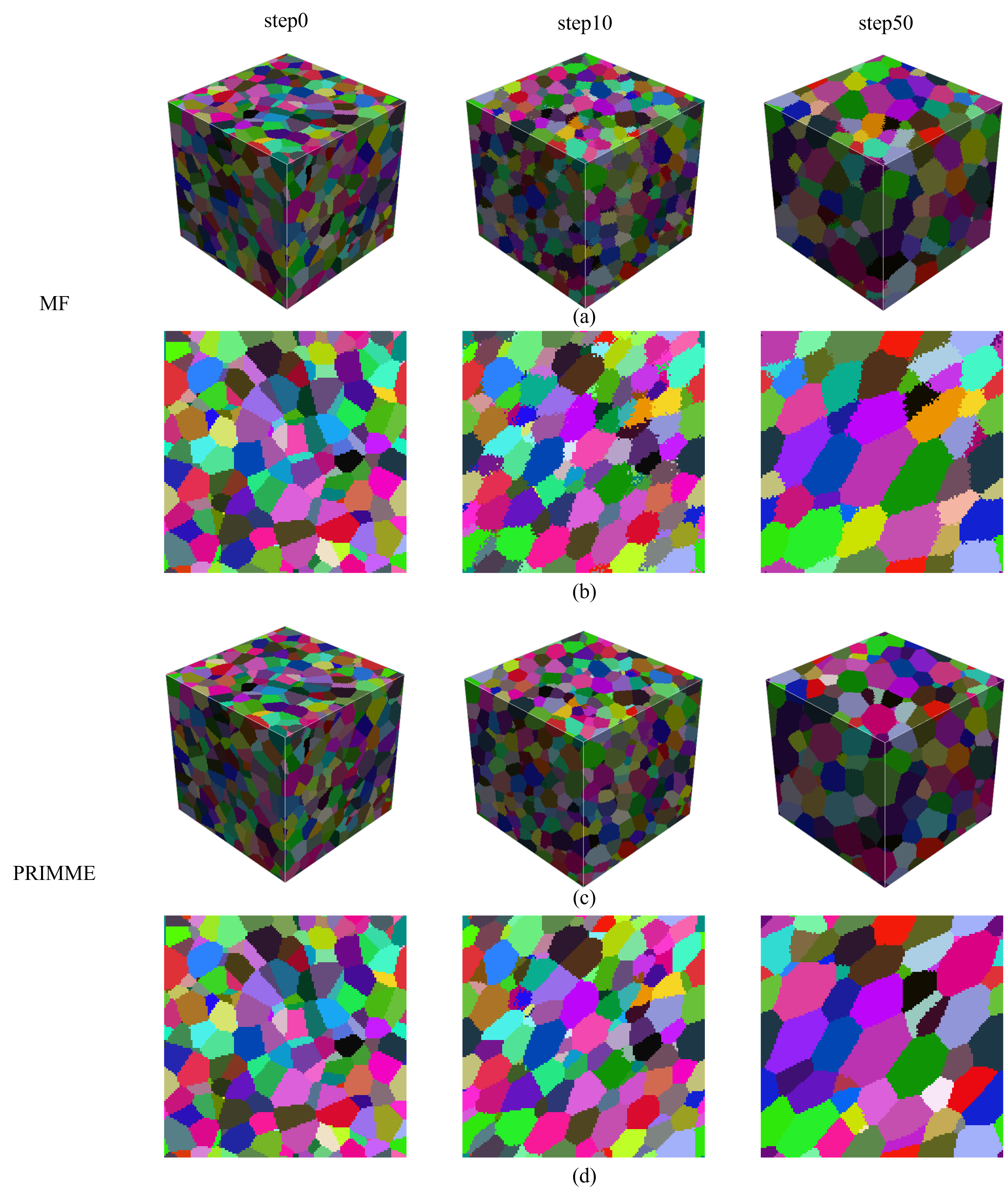}
  \caption{Inclination-dependent grain-growth evolution in MF and 3D-PRIMME. Representative 3D microstructures ((a) and (c)) and corresponding 2D cross-sections ((b) and (d)) are shown for the MF reference ((a) and (b)) and 3D-PRIMME inference ((c) and (d)) at time steps 1, 2, 10, and 50. }
\label{fig:3dprimme_inclination}
\end{figure}

Beyond ideal isotropic grain growth, we further evaluate our model on inclination-dependent datasets. This is important because inclination dependence is a key characteristic of anisotropic grain growth and cannot be described by isotropic coarsening behavior alone.  As shown in Fig.~\ref{fig:3dprimme_inclination}, 3D-PRIMME trained on inclination-dependent mode filter data qualitatively reproduces the resulting inclination-dependent growth behavior.

\begin{figure}[pos=htbp]
  \centering
    \includegraphics[width=0.8\linewidth]{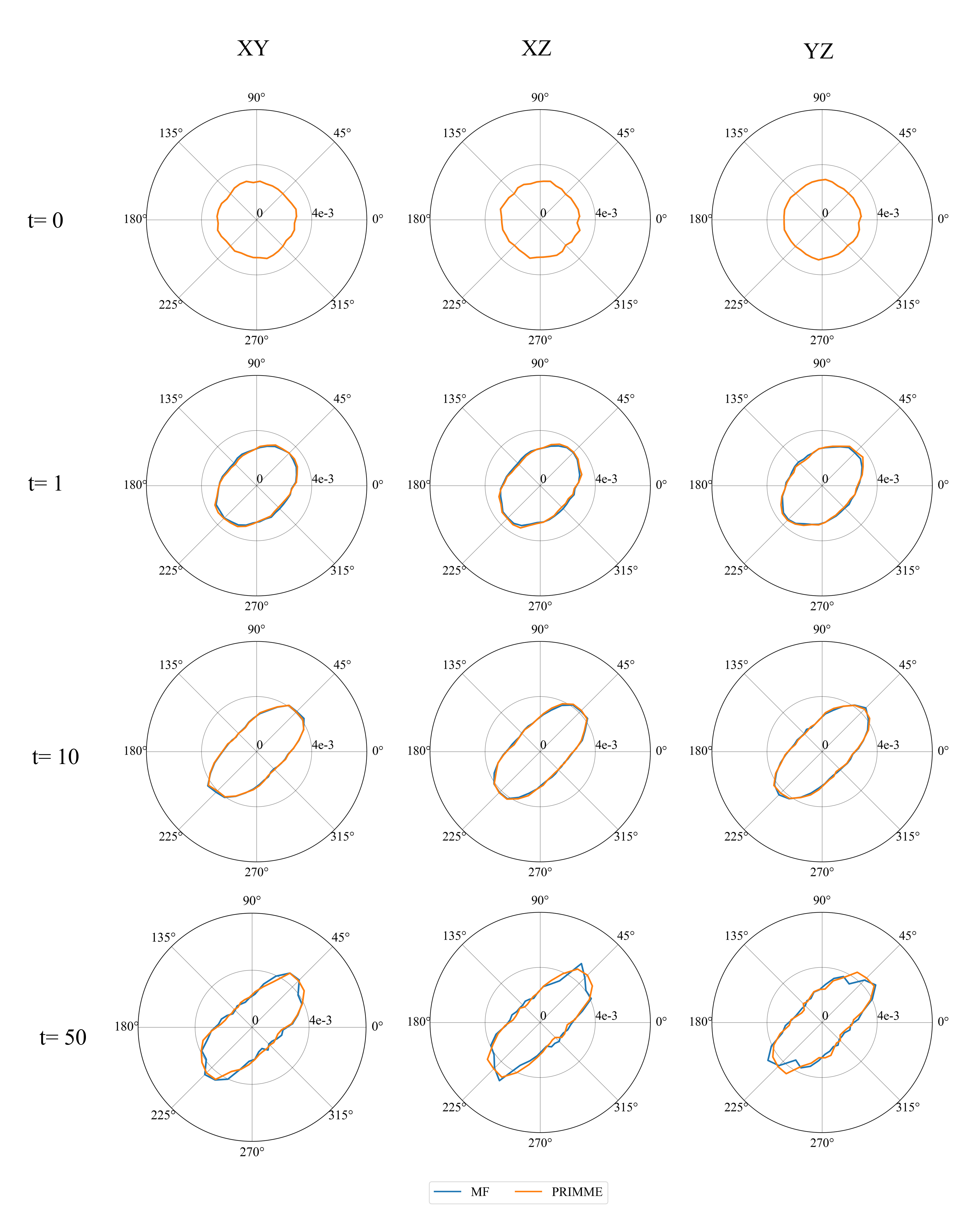}
\caption{Comparison of inclination distributions predicted by MF and 3D-PRIMME. Polar plots of the inclination distributions in the XY, XZ, and YZ planes are shown for time steps 0, 1, 10, and 50. }
\label{fig:inclination_distribution}
\end{figure}

To further examine whether the anisotropy introduced by the Gaussian kernel is preserved during grain growth, we compute the inclination distributions in the three orthogonal projection planes \cite{yang2024triple}. As shown in Fig.~\ref{fig:inclination_distribution}, 3D-PRIMME remains in close agreement with the MF reference across all three projections as the anisotropy develops over time. The consistent evolution of these distributions indicates that 3D-PRIMME is able to learn the inclination-dependent growth behavior from mode filter introduced by the anisotropic kernel. This capability is particularly important for future applications to experimental data, where inclination-dependent effects may be present but are not explicitly prescribed, and therefore represents a step toward learning physically meaningful anisotropic grain-growth behavior directly from observations. Corresponding stereographic projections are provided in  Fig. S3 of the Supporting Information for visual comparison of the full inclination-density evolution \cite{yang2022calculating}.

\section{Discussion} \label{sec:discussion}

The results demonstrate that 3D-PRIMME can reproduce key kinetic, topological, and statistical features of three-dimensional grain growth while extrapolating to larger simulation domains. These findings indicate that the learned local evolution operator remains stable and physically consistent beyond the training setting.

In contrast to the previous 2D-PRIMME formulation, which required an explicit regularization term in the loss function to obtain stable predictions, the present three-dimensional 3D-PRIMME achieves stable grain-growth evolution without this additional term. This suggests that the window-based construction of the interface-site representation provides an implicit constraint on the learned evolution by limiting both the physical context used to encode the local microstructure and the information provided to the neural network. As shown in Fig.~2, the prediction quality depends on both the observation window size, $N_o$, and the action window size, $N_a$, with the observation window playing a particularly important role in constructing an informative interface-site representation. The sensitivity to window size also highlights an important modeling consideration for applying 3D-PRIMME to different microstructural conditions. Window sizes that are too small may not capture sufficient local context for accurate boundary migration, whereas excessively large windows may introduce unnecessary or less relevant information and degrade prediction accuracy. Therefore, although fixed window sizes are effective for the cases studied here, the optimal choices of $N_o$ and $N_a$ may depend on the characteristic length scales and statistical state of the input microstructure. A promising future direction is to develop an adaptive window-selection strategy in which the observation and action window sizes are adjusted according to evolving microstructural statistics, such as the average grain size, grain-size distribution, and number of grains.

The data efficiency of 3D-PRIMME can be understood from two perspectives. First, the model is trained to learn a local update rule rather than to memorize an entire evolution trajectory. A single microstructure already contains a large number of local neighborhoods that provide repeated examples of similar update events. Therefore, the training data are highly redundant: adding more sequences does not necessarily introduce proportionally more independent learning information once the dominant local transition patterns have been sufficiently sampled. In addition, for stochastic datasets such as those generated by the MF, additional sequences may introduce greater variability in local transition behavior, which can make the learning problem more difficult if this variability is not adequately represented or averaged during training. Second, compared with the previous 2D formulation that required 200 sequences of five consecutive steps in the training data, each 3D sequence contains substantially richer spatial information. A 3D microstructure includes many more local neighborhoods, interface geometries, and topological environments than a 2D microstructure of comparable linear size. In this sense, one 3D sequence can provide training information analogous to many 2D slices or local 2D configurations. This increased spatial diversity helps explain why the 3D model can achieve stable predictions with just one or ten sequences of two consecutive steps rather than 200 sequences of five steps required in the 2D case.

Beyond improving data efficiency, learning local update rules also enables 3D-PRIMME to extrapolate beyond the spatial and temporal scales used during training. As shown in Fig.~\ref{fig:spatio temporal}, although the model is trained on relatively small domains and short temporal sequences, the learned update rule can be applied autoregressively to much larger spatial domains and over longer prediction times. This behavior indicates that the model does not simply memorize the global statistics of the training sequences, but instead captures transferable local evolution rules that can be reused across different system sizes and time horizons. In addition, because the prediction at each site depends only on local interface-site representation, the update process is naturally parallelizable and can be implemented efficiently on modern high-performance computing architectures. This locality enables efficient deployment on large 3D microstructures, supporting simulations at spatial scales far beyond those used during training.

The observed spread between different training instances shown in Fig.~\ref{fig:hgyper_tune} can be interpreted as epistemic uncertainty arising from finite data and parameter non-uniqueness~\cite{hullermeier2021aleatoric}. Notably, this variability does not compromise the preservation of global scaling laws or statistical topology, and primarily manifests as long-term trajectory divergence, which is commonly observed in autoregressive surrogate models of dynamical systems because of accumulated stepwise errors during evolution~\cite{mccabe2023towards,list2025differentiability}. Overall, the model demonstrates stable statistical fidelity and robustness with respect to random initialization and data shuffling. To assess robustness with respect to different randomness sources during training, additional models were trained by varying initialization only or shuffling only~\cite{zhuang2022randomness}. The predicted mean trends remained highly consistent across all three settings (Fig.~\ref{fig:hgyper_tune}, Fig. S1 and S2 of the Supporting Information), indicating that the overall grain-growth dynamics are robust to these randomness sources.

The inclination-dependent grain growth results provide further evidence that 3D-PRIMME learns transferable local update rules rather than simply memorizing evolution trajectories. Although the boundary inclination is not provided explicitly as an input feature, the model is able to reproduce the inclination-dependent evolution observed in the MF reference data. Our previous work has demonstrated the ability to model misorientation-dependent grain growth \cite{melville2024anisotropic}, a natural next step is to incorporate an explicit misorientation representation into the present 3D framework. This extension would enable a fully anisotropic 3D-PRIMME model that accounts for both boundary inclination and misorientation effects, thereby providing a path toward applying PRIMME to experimentally measured 3D microstructures and using it as a surrogate model for anisotropic grain-growth evolution.

\section{Conclusion}

In this work, we present 3D-PRIMME, a physics-regulated machine learning framework for modeling 3D grain growth dynamics. The new approach learns a local evolution rule directly from microstructure data, enabling efficient prediction of microstructure evolution without requiring global representations of the system.

Our results demonstrate that the model can learn physically consistent grain growth dynamics using extremely limited training data. Training with only a single sequence containing two consecutive time steps is sufficient to recover the fundamental coarsening behavior, including the linear $\left<r\right>^2$ scaling and consistent topological statistics. Furthermore, the learned dynamics exhibit strong spatiotemporal generalization. A model trained on $100^3$ microstructures can be directly applied to significantly larger domains up to $1024^3$ and maintain stable predictions over long-horizon evolution of 100 time steps. These results indicate that the model captures local curvature-driven grain boundary migration rules that are intrinsically scale-independent. The present framework therefore provides a scalable and data-efficient surrogate for large-scale 3D microstructure evolution. These results indicate that the model captures local curvature-driven grain-boundary migration rules that are transferable across both spatial scales and prediction horizons. In addition, 3D-PRIMME is able to reproduce inclination-dependent grain growth behavior from anisotropic training data, suggesting that the learned local rules can capture direction-dependent boundary migration effects. This capability provides a path toward applying PRIMME to experimentally observed 3D microstructures and realistic anisotropic grain-growth problems.

\section*{Data availability}

The data and code supporting the findings of this study will be made available in a public repository upon publication.

\section*{Declaration of competing interest}

The authors declare that they have no known competing financial interests or personal relationships that could have appeared to influence the work reported in this paper.

\section*{Acknowledgements}
The authors would like to acknowledge financial support by the U.S. Department of Energy, Office
of Science, Basic Energy Sciences under Award \#DE-SC0020384.

\bibliographystyle{model1-num-names}
\bibliography{reference}


\clearpage
\input{SI.tex}
\end{document}

%% file: SI.tex

\clearpage
\appendix

\setcounter{figure}{0}
\setcounter{table}{0}
\setcounter{equation}{0}
\setcounter{page}{1}

\renewcommand{\thefigure}{S\arabic{figure}}
\renewcommand{\thetable}{S\arabic{table}}
\renewcommand{\theequation}{S\arabic{equation}}
\renewcommand{\thepage}{S\arabic{page}}

\section*{Supplementary Information}

\begin{figure}[H]
    \centering
    \includegraphics[width=\columnwidth]{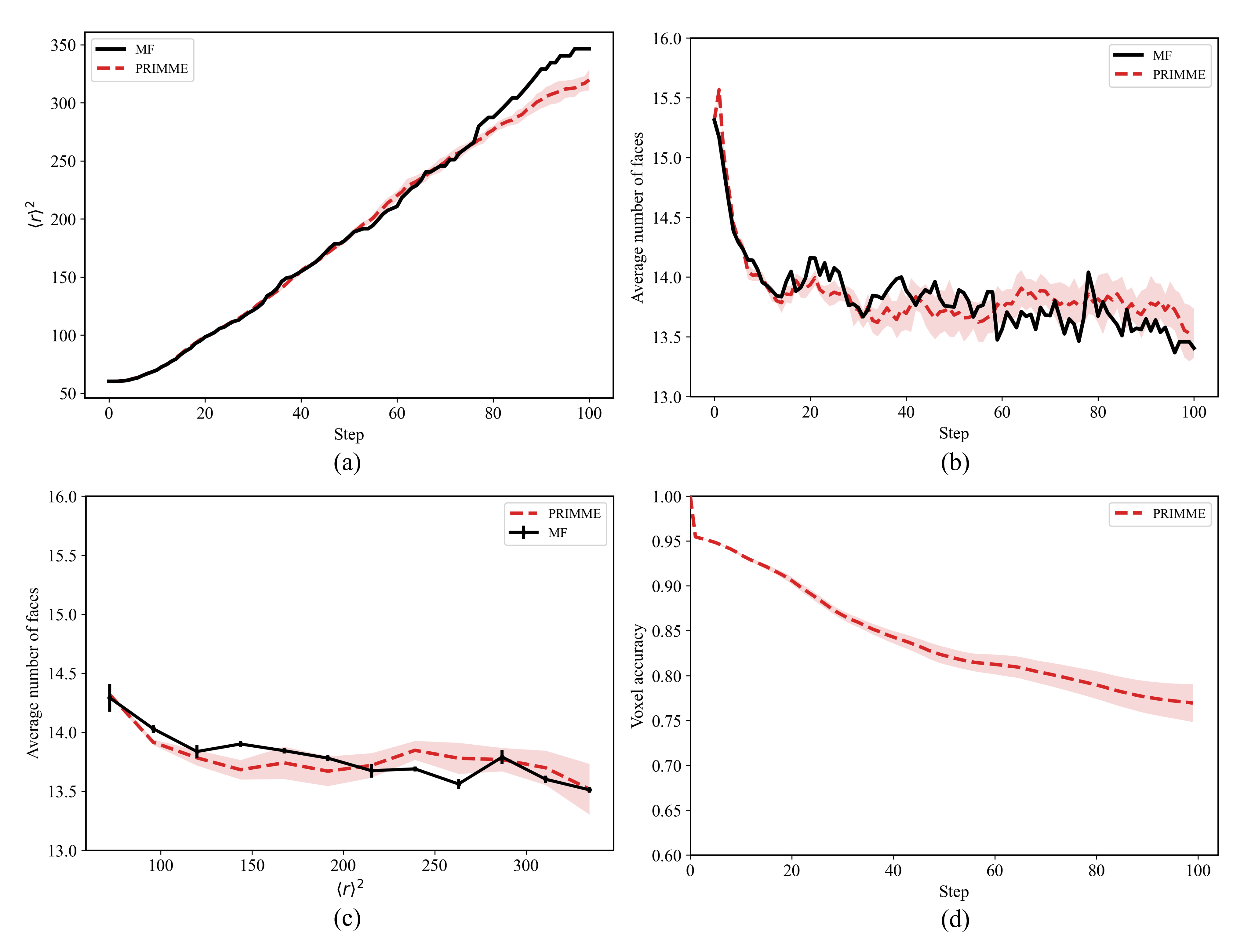}
    \caption{Statistical comparison between MF and 3D-PRIMME predictions with uncertainty quantification when only the random initialization is varied across runs, while the data-shuffling seed is fixed. (a) Mean squared grain size, $\langle r \rangle^2$, as a function of step. (b) Average number of grain faces as a function of step. (c) Average number of grain faces as a function of $\langle r \rangle^2$. (d) Voxel-wise accuracy as a function of step. The solid black curves denote the MF reference data, and the red dashed curves denote the mean 3D-PRIMME predictions. The shaded regions and error bars indicate the corresponding standard deviations across independent runs. The results show that varying the model initialization alone produces only modest variability, while the overall grain-growth kinetics and topological trends remain consistent.}
    \label{fig:uncertainty_only_init}
\end{figure}

\begin{figure}[H]
    \centering
    \includegraphics[width=\columnwidth]{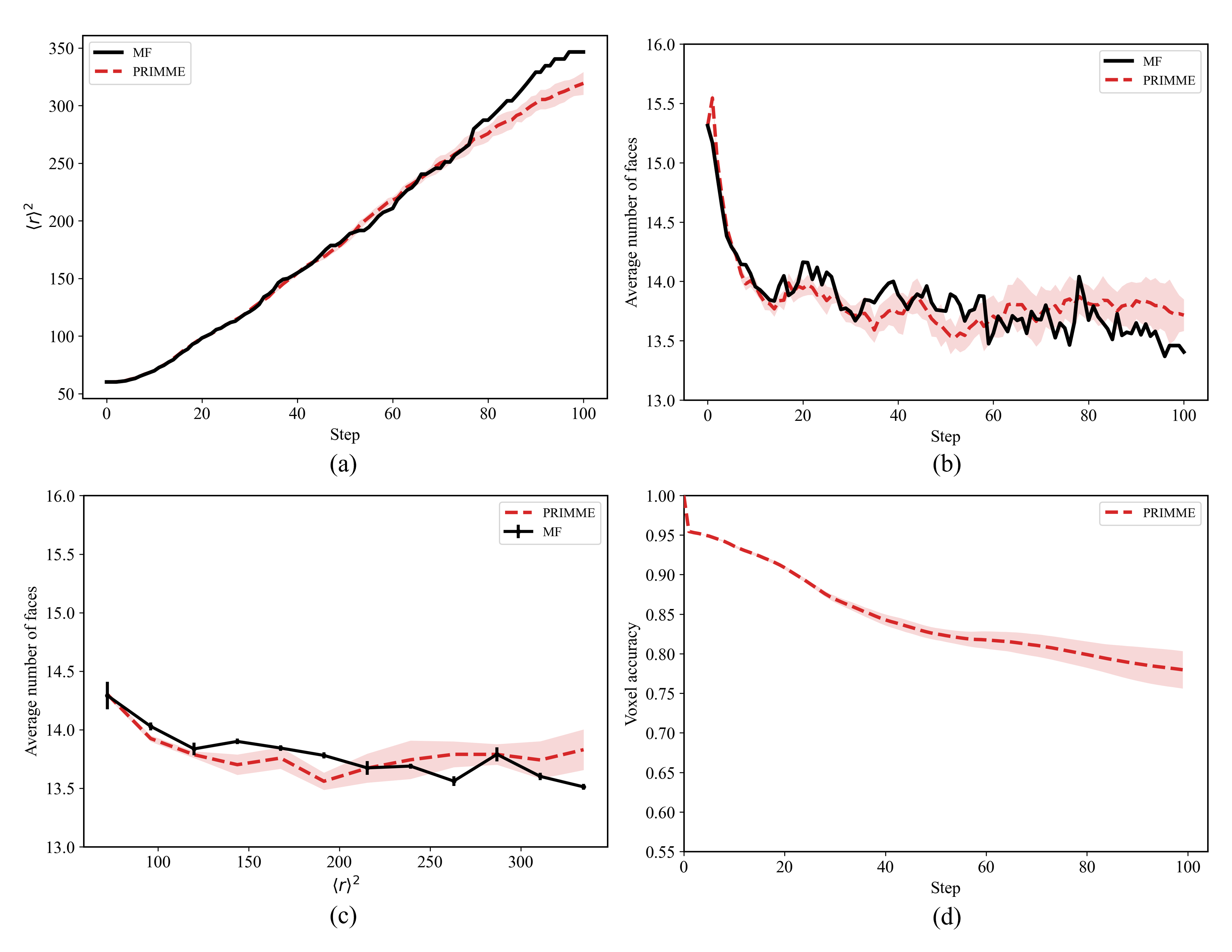}
    \caption{Statistical comparison between MF and 3D-PRIMME predictions with uncertainty quantification when only the data-shuffling seed is varied across runs, while the random initialization is fixed. (a) Mean squared grain size, $\langle r \rangle^2$, as a function of step. (b) Average number of grain faces as a function of step. (c) Average number of grain faces as a function of $\langle r \rangle^2$. (d) Voxel-wise accuracy as a function of step. The solid black curves denote the MF reference data, and the red dashed curves denote the mean 3D-PRIMME predictions. The shaded regions and error bars indicate the corresponding standard deviations across independent runs. The results show that varying the data-shuffling seed alone also leads to limited variability, without altering the overall grain-growth kinetics and topological trends.}
    \label{fig:uncertainty_only_shuffling}
\end{figure}

\begin{table}[pos=htbp]
\centering
\caption{
Quantitative comparison of observation and action window sizes.
The relative error in $\langle r\rangle^2$ and the MAE in the average number of grain faces are computed relative to the MF reference over the 100-step rollout.
The mean accuracy is the time-averaged voxel-wise prediction accuracy.
}
\label{tab:window_size_metrics}
\begin{tabular}{ccccc}
\toprule
{$N_o$} & {$N_a$} 
& {Rel. error in $\langle r\rangle^2$} 
& {MAE in faces} 
& {Mean accuracy} \\
\midrule
7  & 5 & 0.1251 & 0.1492 & 0.8157 \\
7  & 7 & 0.0656 & 0.1498 & 0.8092 \\
7  & 9 & 0.0603 & 0.1948 & 0.8096 \\
\midrule
9  & 5 & 0.0349 & {\bfseries 0.1385} & 0.8468 \\
9  & 7 & 0.0288 & 0.1520 & {\bfseries 0.8484} \\
9  & 9 & {\bfseries 0.0285} & 0.1710 & 0.8432 \\
\midrule
11 & 5 & 0.0921 & 0.1954 & 0.8412 \\
11 & 7 & 0.0609 & 0.1555 & 0.8156 \\
11 & 9 & 0.0471 & 0.1769 & 0.8031 \\
\bottomrule
\end{tabular}
\end{table}

\begin{figure}[H]
    \centering
    \includegraphics[width=\columnwidth]{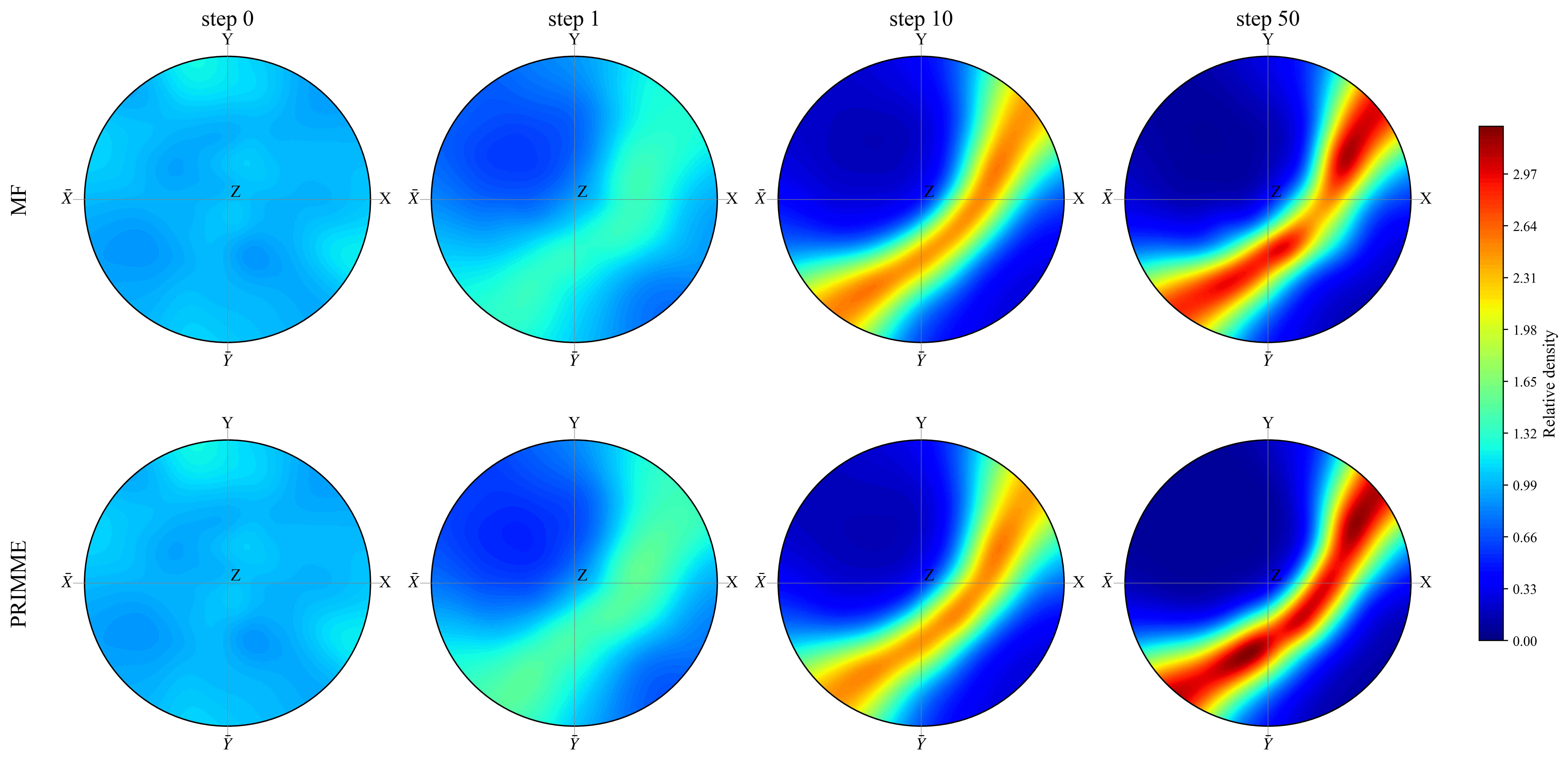}
    \caption{Comparison of stereographic projections of inclination distributions for the MF reference, top row, and 3D-PRIMME prediction, bottom row, at simulation steps 0, 1, 10, and 50. Colors indicate the relative density of interface-normal orientations on the stereographic plane. Both methods evolve from an initially near-uniform distribution toward a pronounced anisotropic band, showing that 3D-PRIMME accurately captures the development and persistence of the inclination-dependent texture observed in MF.}
    \label{fig:stereographic_projection}
\end{figure}

%% file: reference.bib
@article{wu1982potts,
  title={The potts model},
  author={Wu, Fa-Yueh},
  journal={Reviews of modern physics},
  volume={54},
  number={1},
  pages={235},
  year={1982},
  publisher={APS}
}

@article{chen2002phase,
  title={Phase-field models for microstructure evolution},
  author={Chen, Long-Qing},
  journal={Annual review of materials research},
  volume={32},
  number={1},
  pages={113--140},
  year={2002},
  publisher={Annual Reviews 4139 El Camino Way, PO Box 10139, Palo Alto, CA 94303-0139, USA}
}

@article{kamachali20123,
  title={3-D phase-field simulation of grain growth: Topological analysis versus mean-field approximations},
  author={Kamachali, Reza Darvishi and Steinbach, Ingo},
  journal={Acta Materialia},
  volume={60},
  number={6-7},
  pages={2719--2728},
  year={2012},
  publisher={Elsevier}
}

@article{yan2022novel,
  title={A novel physics-regularized interpretable machine learning model for grain growth},
  author={Yan, Weishi and Melville, Joseph and Yadav, Vishal and Everett, Kristien and Yang, Lin and Kesler, Michael S and Krause, Amanda R and Tonks, Michael R and Harley, Joel B},
  journal={Materials \& Design},
  volume={222},
  pages={111032},
  year={2022},
  publisher={Elsevier}
}

@article{fan2024accelerate,
  title={Accelerate microstructure evolution simulation using graph neural networks with adaptive spatiotemporal resolution},
  author={Fan, Shaoxun and Hitt, Andrew L and Tang, Ming and Sadigh, Babak and Zhou, Fei},
  journal={Machine Learning: Science and Technology},
  volume={5},
  number={2},
  pages={025027},
  year={2024},
  publisher={IOP Publishing}
}

@article{yang2021self,
  title={Self-supervised learning and prediction of microstructure evolution with convolutional recurrent neural networks},
  author={Yang, Kaiqi and Cao, Yifan and Zhang, Youtian and Fan, Shaoxun and Tang, Ming and Aberg, Daniel and Sadigh, Babak and Zhou, Fei},
  journal={Patterns},
  volume={2},
  number={5},
  year={2021},
  publisher={Elsevier}
}

@article{peivaste2025teaching,
  title={Teaching artificial intelligence to perform rapid, resolution-invariant grain growth modeling via Fourier Neural Operator},
  author={Peivaste, Iman and Makradi, Ahmed and Belouettar, Salim},
  journal={Computer Methods in Applied Mechanics and Engineering},
  volume={440},
  pages={117945},
  year={2025},
  publisher={Elsevier}
}

@article{tep2025high,
  title={High-fidelity Grain Growth Modeling: Leveraging Deep Learning for Fast Computations},
  author={Tep, Pungponhavoan and Bernacki, Marc},
  journal={arXiv preprint arXiv:2505.05354},
  year={2025}
}

@article{tian2025scaling,
  title={Scaling Kinetic Monte-Carlo Simulations of Grain Growth with Combined Convolutional and Graph Neural Networks},
  author={Tian, Zhihui and Suwandi, Ethan and Oppelstrup, Tomas and Bulatov, Vasily V and Harley, Joel B and Zhou, Fei},
  journal={arXiv preprint arXiv:2511.17848},
  year={2025}
}

@article{melville2024new,
  title={A new efficient grain growth model using a random Gaussian-sampled mode filter},
  author={Melville, Joseph and Yadav, Vishal and Yang, Lin and Krause, Amanda R and Tonks, Michael R and Harley, Joel B},
  journal={Materials \& Design},
  volume={237},
  pages={112604},
  year={2024},
  publisher={Elsevier}
}

@article{hestness2017deep,
  title={Deep learning scaling is predictable, empirically},
  author={Hestness, Joel and Narang, Sharan and Ardalani, Newsha and Diamos, Gregory and Jun, Heewoo and Kianinejad, Hassan and Patwary, Md Mostofa Ali and Yang, Yang and Zhou, Yanqi},
  journal={arXiv preprint arXiv:1712.00409},
  year={2017}
}

@article{hullermeier2021aleatoric,
  title={Aleatoric and epistemic uncertainty in machine learning: An introduction to concepts and methods},
  author={H{\"u}llermeier, Eyke and Waegeman, Willem},
  journal={Machine learning},
  volume={110},
  number={3},
  pages={457--506},
  year={2021},
  publisher={Springer}
}

@article{wakai2000three,
  title={Three-dimensional microstructural evolution in ideal grain growth—general statistics},
  author={Wakai, Fumihiro and Enomoto, Naoya and Ogawa, Hiroshi},
  journal={Acta Materialia},
  volume={48},
  number={6},
  pages={1297--1311},
  year={2000},
  publisher={Elsevier}
}

@article{miodownik2002review,
  title={A review of microstructural computer models used to simulate grain growth and recrystallisation in aluminium alloys},
  author={Miodownik, Mark A},
  journal={Journal of Light Metals},
  volume={2},
  number={3},
  pages={125--135},
  year={2002},
  publisher={Elsevier}
}

@article{ludwig2009new,
  title={New opportunities for 3D materials science of polycrystalline materials at the micrometre lengthscale by combined use of X-ray diffraction and X-ray imaging},
  author={Ludwig, Wolfgang and King, A and Reischig, P and Herbig, M and Lauridsen, Erik Mejdal and Schmidt, S{\o}ren and Proudhon, Henry and Forest, Samuel and Cloetens, Peter and Du Roscoat, S Rolland and others},
  journal={Materials Science and Engineering: A},
  volume={524},
  number={1-2},
  pages={69--76},
  year={2009},
  publisher={Elsevier}
}

@article{holm2006three,
  title={Three-dimensional materials science},
  author={Holm, Elizabeth A and Duxbury, Phillip M},
  journal={Scripta materialia},
  volume={54},
  number={6},
  pages={1035--1040},
  year={2006},
  publisher={Elsevier}
}

@article{yamakov2006relation,
  title={Relation between grain growth and grain-boundary diffusion in a pure material by molecular dynamics simulations},
  author={Yamakov, V and Moldovan, D and Rastogi, K and Wolf, D},
  journal={Acta materialia},
  volume={54},
  number={15},
  pages={4053--4061},
  year={2006},
  publisher={Elsevier}
}

@article{xu2024grain,
  title={Grain boundary migration in polycrystalline $\alpha$-Fe},
  author={Xu, Zipeng and Shen, Yu-Feng and Naghibzadeh, S Kiana and Peng, Xiaoyao and Muralikrishnan, Vivekanand and Maddali, Siddharth and Menasche, David and Krause, Amanda R and Dayal, Kaushik and Suter, Robert M and others},
  journal={Acta Materialia},
  volume={264},
  pages={119541},
  year={2024},
  publisher={Elsevier}
}

@article{mccabe2023towards,
  title={Towards stability of autoregressive neural operators},
  author={McCabe, Michael and Harrington, Peter and Subramanian, Shashank and Brown, Jed},
  journal={arXiv preprint arXiv:2306.10619},
  year={2023}
}

@article{list2025differentiability,
  title={Differentiability in unrolled training of neural physics simulators on transient dynamics},
  author={List, Bjoern and Chen, Li-Wei and Bali, Kartik and Thuerey, Nils},
  journal={Computer Methods in Applied Mechanics and Engineering},
  volume={433},
  pages={117441},
  year={2025},
  publisher={Elsevier}
}

@article{yang2022calculating,
  title={Calculating the grain boundary inclination of voxelated grain structures using a smoothing algorithm},
  author={Yang, Lin and Hilty, Floyd and Muralikrishnan, Vivekanand and Silva-Reyes, Kenneth and Harley, Joel B and Krause, Amanda R and Tonks, Michael R},
  journal={Scripta Materialia},
  volume={218},
  pages={114796},
  year={2022},
  publisher={Elsevier}
}

@article{yang2024triple,
  title={A triple junction energy study using an inclination-dependent anisotropic Monte Carlo Potts grain growth model},
  author={Yang, Lin and Yadav, Vishal and Melville, Joseph and Harley, Joel B and Krause, Amanda R and Tonks, Michael R},
  journal={Materials \& Design},
  volume={239},
  pages={112763},
  year={2024},
  publisher={Elsevier}
}

@article{miyoshi2017ultra,
  title={Ultra-large-scale phase-field simulation study of ideal grain growth},
  author={Miyoshi, Eisuke and Takaki, Tomohiro and Ohno, Munekazu and Shibuta, Yasushi and Sakane, Shinji and Shimokawabe, Takashi and Aoki, Takayuki},
  journal={NPJ Computational Materials},
  volume={3},
  number={1},
  pages={25},
  year={2017},
  publisher={Nature Publishing Group UK London}
}

@article{zhuang2022randomness,
  title={Randomness in neural network training: Characterizing the impact of tooling},
  author={Zhuang, Donglin and Zhang, Xingyao and Song, Shuaiwen and Hooker, Sara},
  journal={Proceedings of Machine Learning and Systems},
  volume={4},
  pages={316--336},
  year={2022}
}

@article{mason2015kinetics,
  title={Kinetics and anisotropy of the Monte Carlo model of grain growth},
  author={Mason, JK and Lind, J and Li, SF and Reed, BW and Kumar, M},
  journal={Acta Materialia},
  volume={82},
  pages={155--166},
  year={2015},
  publisher={Elsevier}
}

@inproceedings{mindermann2022prioritized,
  title={Prioritized training on points that are learnable, worth learning, and not yet learnt},
  author={Mindermann, S{\"o}ren and Brauner, Jan M and Razzak, Muhammed T and Sharma, Mrinank and Kirsch, Andreas and Xu, Winnie and H{\"o}ltgen, Benedikt and Gomez, Aidan N and Morisot, Adrien and Farquhar, Sebastian and others},
  booktitle={International Conference on Machine Learning},
  pages={15630--15649},
  year={2022},
  organization={PMLR}
}

@article{anderson1984computer,
  title={Computer simulation of grain growth—I. Kinetics},
  author={Anderson, MP and Srolovitz, DJ and Grest, GS and Sahni, PS},
  journal={Acta metallurgica},
  volume={32},
  number={5},
  pages={783--791},
  year={1984},
  publisher={Elsevier}
}

@article{he2006computer,
  title={Computer simulation of 2D grain growth using a cellular automata model based on the lowest energy principle},
  author={He, Yizhu and Ding, Hanlin and Liu, Liufa and Shin, Keesam},
  journal={Materials Science and Engineering: A},
  volume={429},
  number={1-2},
  pages={236--246},
  year={2006},
  publisher={Elsevier}
}

@article{lazar2011more,
  title={A more accurate three-dimensional grain growth algorithm},
  author={Lazar, Emanuel A and Mason, Jeremy K and MacPherson, Robert D and Srolovitz, David J},
  journal={Acta Materialia},
  volume={59},
  number={17},
  pages={6837--6847},
  year={2011},
  publisher={Elsevier}
}

@article{fausty2021new,
  title={A new analytical test case for anisotropic grain growth problems},
  author={Fausty, Julien and Murgas, Brayan and Florez, Sebastian and Bozzolo, Nathalie and Bernacki, Marc},
  journal={Applied Mathematical Modelling},
  volume={93},
  pages={28--52},
  year={2021},
  publisher={Elsevier}
}

@article{melville2024anisotropic,
  title={Anisotropic physics-regularized interpretable machine learning of microstructure evolution},
  author={Melville, Joseph and Yadav, Vishal and Yang, Lin and Krause, Amanda R and Tonks, Michael R and Harley, Joel B},
  journal={Computational Materials Science},
  volume={238},
  pages={112941},
  year={2024},
  publisher={Elsevier}
}

@article{lyu2026limits,
  title={On the Limits of Predictivity in Microstructural Evolution Simulations: Ensemble Simulations of Polycrystalline Grain Growth},
  author={Lyu, Meizhong and Holm, Elizabeth A},
  journal={Metallurgical and Materials Transactions A},
  pages={1--14},
  year={2026},
  publisher={Springer}
}
